\documentclass[letterpaper, 10 pt, conference]{ieeeconf}  
\IEEEoverridecommandlockouts                              
\usepackage{ps}

\preprinttrue
\publishedtrue

\glsdisablehyper 
\newcommand{\anAgent}{\ensuremath{i}}
\newcommand{\anotherAgent}{\ensuremath{j}}
\newcommand{\timestep}{\ensuremath{k}}
\newcommand{\timeStep}{\ensuremath{t}}

\newcommand{\ofATimeStep}{\ensuremath{_{\timeStep}}}
\newcommand{\ofNextTimeStep}{\ensuremath{_{\timeStep+1}}}

\newcommand{\ofAnAgent}{\ensuremath{^{(\anAgent)}}}
\newcommand{\ofAnotherAgent}{\ensuremath{^{(\anotherAgent)}}}
\newcommand{\ofAgent}[1]{\ensuremath{^{(#1)}}}

\newglossaryentry{matrix:Adjacency}{
	name=\ensuremath{\bm{D}},
	description={Adjacency matrix},
	sort={D},
    type=symbol
}
\newcommand{\matAdjacency}{\gls{matrix:Adjacency}}
\newcommand{\matAdjacencyElement}[1]{\glslink{matrix:Adjacency}{\matAdjacency_{#1}}}

\newglossaryentry{set:realNumbers}{
	name=\ensuremath{\mathbb{R}},
	description={Set of real numbers},
	sort={real numbers},
    type=symbol
}
\newcommand{\setRealNumbers}{\gls{set:realNumbers}}

\newglossaryentry{set:naturalNumbers}{
	name=\ensuremath{\mathbb{N}},
	description={Set of natural numbers},
	sort={natural numbers},
    type=symbol
}

\newglossaryentry{set:systemStates}{
	name=\ensuremath{\mathcal{S}},
	description={Set of system states},
	sort={set of system states},
    type=symbol
}

\newglossaryentry{set:bigO}{
	name=\ensuremath{O},
	description={Big O},
	sort={O},
    type=symbol
}

\newglossaryentry{scalar:Weight}{
	name=\ensuremath{w},
	description={Weight},
	sort={weight},
    type=symbol
}

\newglossaryentry{scalar:NumberOfAgents}{
    name=\ensuremath{N_A},
    description={Number of agents},
    sort={Number of agents},
    type=symbol
}
\newcommand{\numAgents}{\gls{scalar:NumberOfAgents}}

\newglossaryentry{graph:path}{
    name=\ensuremath{\pi},
    description={Path},
    sort={Path},
    type=symbol
}
\newcommand{\graphPath}{\gls{graph:path}}

\newglossaryentry{scalar:NumberOfVerticesInPath}{
    name=\ensuremath{N_{\graphPath}},
    description={Number of vertices in path $\graphPath$, or length},
    sort={Number of vertices in path},
    type=symbol
}
\newcommand{\numVerticesPath}{\gls{scalar:NumberOfVerticesInPath}}

\newglossaryentry{trajectory:Reference}{
    name=\ensuremath{\bm{r}},
    description={Reference trajectory},
    sort={Reference Trajectory},
    type=symbol
}

\newglossaryentry{sym:horizonControl}{
	name=\ensuremath{N_u},
	description={Control horizon in model predictive control},
	sort={Nu},
    type=symbol
}

\newglossaryentry{sym:horizonPrediction}{
	name=\ensuremath{N_p},
	description={Prediction horizon in model predictive control},
	sort={Np},
    type=symbol
}

\newglossaryentry{sym:vehicleOrientation}{
	name=\ensuremath{\psi},
	description={Vehicle orientation},
	sort={psi},
    type=symbol
}

\newglossaryentry{sym:sysModelContinuous}{
    name=\ensuremath{f},
    description={Continuous-time system model},
    sort={f continuous-time},
    type=symbol
}
\newcommand{\sysModelContinuous}{\gls{sym:sysModelContinuous}}

\newglossaryentry{sym:sysModelDiscrete}{
    name=\ensuremath{f_{d}},
    description={Discrete-time system model},
    sort={f discrete-time},
    type=symbol
}

\newglossaryentry{sym:sysControlInputs}{
	name=\ensuremath{\bm{u}},
	description={System control inputs},
	sort=u,
    type=symbol
}
\NewDocumentCommand{\sysControlInputs}{ o }{\glslink{sym:sysControlInputs}{%
    \IfNoValueTF{#1}%
        {\ensuremath{\bm{u}}}%
        {\ensuremath{\bm{u}^{(#1)}}}%
}}

\newglossaryentry{sym:outputs}{
	name=\ensuremath{\bm{y}},
	description={System outputs},
	sort={y},
    type=symbol
}

\newglossaryentry{sym:sysSpeed}{
	name=\ensuremath{\mathrm{v}},
	description={Vehicle speed},
	sort={v},
    type=symbol
}
\newcommand{\sysSpeed}{\gls{sym:sysSpeed}}

\newglossaryentry{sym:inSpeed}{
	name=\ensuremath{u_{\sysSpeed}},
	description={Vehicle input speed},
	sort={uv},
    type=symbol
}

\newglossaryentry{sym:steering-angle}{
	name=\ensuremath{\delta},
	description={Vehicle steering angle},
	sort={delta},
    type=symbol
}

\newglossaryentry{sym:inSteering}{
	name=\ensuremath{u_{\delta}},
	description={Vehicle input steering angle},
	sort={ud},
    type=symbol
}

\newglossaryentry{sym:nColors}{
	name=\ensuremath{N_c},
	description={Number of colors},
	sort={Number of colors},
    type=symbol
}

\newglossaryentry{sym:nStates}{
	name=\ensuremath{n},
	description={Number of states of a dynamical system},
	sort={Number of states},
    type=symbol
}
\newcommand{\numStates}{\gls{sym:nStates}}

\newglossaryentry{sym:nInputs}{
    name=\ensuremath{m},
    description={Number of inputs of a dynamical system},
    sort={m number of inputs},
    type=symbol
}
\newcommand{\numInputs}{\gls{sym:nInputs}}

\newglossaryentry{sym:nLevels}{
	name=\ensuremath{N_{\text{CL}}},
	description={Number of computation levels},
	sort={Number of computation levels},
    type=symbol
}

\newglossaryentry{sym:nLevelsAllowed}{
	name=\ensuremath{N_{\text{CL},a}},
	description={Allowed number of computation levels},
	sort={Number of computation levels allowed},
    type=symbol
}

\newglossaryentry{sym:numGroups}{
	name=\ensuremath{N_{g}},
	description={Number of parallelly computing groups of agents},
	sort={Number of groups},
    type=symbol
}

\newglossaryentry{sym:fnPrio}{
    name=\ensuremath{p},
    description={Priority assignment function},
    sort={Priority assignment function},
    type=symbol
}

\newglossaryentry{sym:tSample}{
	name=\ensuremath{T_s},
	description={Sample Time},
	sort={T sample},
    type=symbol
}

\newglossaryentry{sym:tSolve}{
	name=\ensuremath{T_\text{sol.}},
	description={Computation time \tSolveB{\anAgent} that agent $\anAgent$ needs to solve its \ac{ocp}},
	sort={T solve},
    type=symbol
}

\newcommand{\tSolveB}[1]{\glslink{sym:tSolve}{\ensuremath{\ensuremath{T_\text{sol.}}^{(#1)}}}}

\newglossaryentry{sym:tSolveUpper}{
	name=\ensuremath{T_\text{sol.,max}},
	description={Upper computation time $T_\text{sol.,max}\ofAgent{\anAgent}$ that agent $\anAgent$ needs to solve it \ac{ocp}},
	sort={T solve upper},
    type=symbol
}

\newglossaryentry{sym:vertices}{
	name=\ensuremath{\mathcal{V}},
	description={Set of vertices},
	sort={Vertices},
    type=symbol
}
\newcommand{\setVertices}{\gls{sym:vertices}}
\newcommand{\setAgents}{\setVertices}

\newcommand{\helpSetPredecessors}[1]{\ensuremath{\setVertices^{(#1\leftarrow)}}}
\newglossaryentry{sym:predecessors}{
	name=\ensuremath{\helpSetPredecessors{i}},
	description={Set of predecessors of vertex $i$},
	sort={Vertices 1},
    type=symbol
}
\newcommand{\setPredecessors}[1]{\glslink{sym:predecessors}{\ensuremath{\helpSetPredecessors{#1}}}}

\newcommand{\helpSetPredecessorsPar}[1]{\ensuremath{\setVertices^{(#1\leftarrow)}_{\text{par.}}}}
\newglossaryentry{sym:predecessorsPar}{
	name=\ensuremath{\helpSetPredecessorsPar{i}},
	description={Set of predecessors of vertex $i$ that have parallel couplings with it},
	sort={Vertices 2},
    type=symbol
}

\newcommand{\helpSetPredecessorsSeq}[1]{\ensuremath{\setVertices^{(#1\leftarrow)}_{\text{seq.}}}}
\newglossaryentry{sym:predecessorsSeq}{
	name=\ensuremath{\helpSetPredecessorsSeq{i}},
	description={Set of predecessors of vertex $i$ that have sequential couplings with it},
	sort={Vertices 3},
    type=symbol
}

\newcommand{\helpSetSuccessors}[1]{\ensuremath{\setVertices^{(#1\rightarrow)}}}
\newglossaryentry{sym:successors}{
	name=\ensuremath{\helpSetSuccessors{i}},
	description={Set of successors of vertex $i$},
	sort={Vertices 4},
    type=symbol
}
\newcommand{\setSuccessors}[1]{\glslink{sym:successors}{\ensuremath{\helpSetSuccessors{#1}}}}

\newglossaryentry{sym:neighbors}{
	name=\ensuremath{\setVertices^{(i)}},
	description={Set of neighbors of vertex $i$},
	sort={Vertices 0},
    type=symbol
}
\newcommand{\setNeighbors}[1]{\glslink{sym:neighbors}{\ensuremath{\setVertices^{(#1)}}}}

\newglossaryentry{sym:degree}{
	name=\ensuremath{d^{(i)}},
	description={Degree of vertex $i$. Sum of in-degree and out-degree},
	sort=degree,
    type=symbol
}
\newcommand{\vertexDegree}[1]{\glslink{sym:degree}{\ensuremath{d^{(#1)}}}}

\newcommand{\helpVertexInDegree}[1]{\ensuremath{d^{(#1\leftarrow)}}}
\newglossaryentry{sym:inDegree}{
    name=\helpVertexInDegree{i},
    description={In-degree of vertex $i$},
    sort={degree in},
    type=symbol
}
\newcommand{\vertexInDegree}[1]{\glslink{sym:inDegree}{\helpVertexInDegree{#1}}}

\newcommand{\helpVertexOutDegree}[1]{\ensuremath{d^{(#1\rightarrow)}}}
\newglossaryentry{sym:outDegree}{
    name=\helpVertexOutDegree{i},
    description={Out-degree of vertex $i$},
    sort={degree out},
    type=symbol,
}
\newcommand{\vertexOutDegree}[1]{\glslink{sym:outDegree}{\helpVertexOutDegree{#1}}}

\newglossaryentry{sym:matLevels}{
	name=\ensuremath{\bm{L}},
	description={Matrix of computation levels},
	sort=L,
    type=symbol
}

\newglossaryentry{sym:tComp}{
	name=\ensuremath{T},
	description={Computation time},
	sort={T},
    type=symbol
}

\newglossaryentry{sym:tCompNcs}{
	name=\ensuremath{T_{\text{NCS}}},
	description={Computation time of \iac{ncs}},
	sort={T NCS},
    type=symbol
}
\newcommand{\tCompNcs}{\gls{sym:tCompNcs}}

\newglossaryentry{graph:Undirected}{
	name=\ensuremath{\mathcal{G}},
	description={Undirected Graph},
	sort={graph1},
    type=symbol
}
\newcommand{\graphUndirected}{\gls{graph:Undirected}}

\newglossaryentry{graph:Directed}{
    name=\ensuremath{\vec{\gls*{graph:Undirected}}},
	description={Directed Graph},
	sort={graph2},
    type=symbol
}
\newcommand{\graphDirected}{\gls{graph:Directed}}

\newglossaryentry{mat:edgeUtilities}{
    name=\ensuremath{M_\text{u}},
	description={Edge utility matrix},
	sort={matrix edge utilities},
    type=symbol
}

\newglossaryentry{sym:setColors}{
    name=\ensuremath{\mathcal{C}},
    description={Set of colors},
    sort=Colors,
    type=symbol
}

\newglossaryentry{sym:varControlInvariantSet}{
	name=\ensuremath{\mathcal{C}_{\text{inv}}},
	description={Control invariant set},
	sort={Control invariant set},
    type=symbol
}

\newglossaryentry{set:Weights}{
	name=\ensuremath{\mathcal{W}},
	description={Set of weights in a weighted graph},
    sort={Weights},
    type=symbol
}

\newglossaryentry{set:Edges}{
	name=\ensuremath{\mathcal{E}},
	description={Set of edges; used to indicate that only undirected edges exist},
    sort={Edges},
    type=symbol
}
\newcommand{\setEdges}{\gls{set:Edges}}

\newglossaryentry{sym:setEdgesDirected}{
	name=\ensuremath{\vec{\gls*{set:Edges}}},
	description={Set of directed edges},
	sort={Edges directed},
    type=symbol
}

\newglossaryentry{sym:varEdge}{
	name=\ensuremath{(i \rightarrow j)},
	description={Directed edge from vertex $i$ to vertex $j$},
	sort={edge},
    type=symbol
}
\newcommand{\edgeDirected}[2]{\glslink{sym:varEdge}{\ensuremath{(#1 \rightarrow #2)}}}

\newglossaryentry{sym:fnReorder}{
	name=\ensuremath{f_r},
	description={Reordering function for graph color values},
	sort=fr,
    type=symbol
}

\newglossaryentry{sym:fcnObjective}{
    name=\ensuremath{J},
    description={Objective function of an optimization problem},
    sort=J,
    type=symbol
}
\NewDocumentCommand{\fcnObjective}{ o }{\glslink{sym:fcnObjective}{%
    \IfNoValueTF{#1}%
        {\ensuremath{J}}%
        {\ensuremath{J^{(#1)}}}%
}}

\newglossaryentry{sym:fcnObjectiveState}{
    name=\ensuremath{\ell_{x}},
    description={Reference tracking objective function},
    sort={lx Reference tracking objective function},
    type=symbol
}
\NewDocumentCommand{\fcnObjectiveState}{ o }{\glslink{sym:fcnObjectiveState}{%
    \IfNoValueTF{#1}%
        {\ensuremath{\ell_{x}}}%
        {\ensuremath{\ell_{x}^{(#1)}}}%
}}

\newglossaryentry{sym:fcnObjectiveStateTerminal}{
    name=\ensuremath{\ell_{f}},
    description={Reference tracking objective terminal function},
    sort={lf Reference tracking objective terminal function},
    type=symbol
}

\newglossaryentry{sym:fcnObjectiveInput}{
    name=\ensuremath{\ell_{u}},
    description={Input change objective function},
    sort={lu Input change objective function},
    type=symbol
}
\NewDocumentCommand{\fcnObjectiveInput}{ o }{\glslink{sym:fcnObjectiveInput}{%
    \IfNoValueTF{#1}%
        {\ensuremath{\ell_{u}}}%
        {\ensuremath{\ell_{u}^{(#1)}}}%
}}

\newglossaryentry{sym:fcnObjectiveCoupling}{
    name=\ensuremath{\ell_\text{c}},
    description={Coupling objective function},
    sort={lc Coupling objective function},
    type=symbol
}
\NewDocumentCommand{\fcnObjectiveCoupling}{ oo }{\glslink{sym:fcnObjectiveCoupling}{%
    \IfNoValueTF{#1}%
        {\ensuremath{\ell_\text{c}}}%
        {\ensuremath{\ell_\text{c}^{(#1,#2)}}}%
}}

\newglossaryentry{sym:fcnConstraintCoupling}{
    name=\ensuremath{c_\text{c}},
    description={Coupling constraint function},
    sort={cc Coupling constraint function},
    type=symbol
}
\NewDocumentCommand{\fcnConstraintCoupling}{ oo }{\glslink{sym:fcnConstraintCoupling}{%
    \IfNoValueTF{#1}%
        {\ensuremath{c_\text{c}}}%
        {\ensuremath{c_\text{c}^{(#1,#2)}}}%
}}

\newglossaryentry{sym:prediction}{
	name=\ensuremath{\tilde{\bm{x}}^{(j \leftarrow i)}_{\cdot \vert k}},
	description={Prediction in agent $i$ for agent $j$ at time $k$},
	sort=x,
    type=symbol,
}
\newcommand{\agentPrediction}{\glslink{sym:prediction}{\ensuremath{ \tilde{\bm{x}} }}}

\newcommand{\agentPredictionForAgentAFromAgentBAtTimeC}[3]{\glslink{sym:prediction}{\ensuremath{ \agentPrediction^{(#1 \leftarrow #2)}_{\cdot \vert #3} }}}

\newglossaryentry{sym:state}{
	name=\ensuremath{\bm{x}},
	description={System state},
	sort=x,
    type=symbol
}
\newcommand{\sysState}{\gls{sym:state}}
\newglossaryentry{sym:ref}{
	name=\ensuremath{\bm{r}},
	description={System reference},
	sort=r,
    type=symbol
}

\newglossaryentry{sym:stateAgent}{
	name=\ensuremath{\sysState^{(i)}_{(k)}},
	description={System state of agent $i$ at time $k$},
	sort=x,
    type=symbol,
}

\newglossaryentry{sym:setReachable}{
	name=\ensuremath{\mathcal{R}^{(i)}},
	description={reachable set of agent $i$},
	sort={Reachable set},
    type=symbol
}
\newcommand{\setReachable}{\glslink{sym:setReachable}{\ensuremath{\mathcal{R}}}}

\newglossaryentry{set:occupiedArea}{
	name=\ensuremath{\mathcal{O}^{(i)}},
	description={Set of the occupied area of the predicted trajectory of agent $\anAgent$},
	sort={occupied area},
    type=symbol
}

\newglossaryentry{set:feasibleStates}{
	name=\ensuremath{\mathcal{X}},
	description={set of feasible states},
	sort={x},
    type=symbol
}
\newcommand{\setFeasibleStates}{\gls{set:feasibleStates}}

\newglossaryentry{set:feasibleInputs}{
	name=\ensuremath{\mathcal{U}},
	description={set of feasible inputs},
	sort={u},
    type=symbol
}
\newcommand{\setFeasibleInputs}{\gls{set:feasibleInputs}}

\newglossaryentry{sym:numStatesConfSpace}{
    name=\ensuremath{n_p},
    description={Number of states that are in the conflictual space},
    sort={n number of states that are in the conflictual space},
    type=symbol
}

\newglossaryentry{sym:fcnProj}{
    name=\text{proj},
    description={A function that projects a reachable set of system states in the conflictual space},
    sort={Project function},
    type=symbol
}

\newglossaryentry{rl:setOfAgents}{
    name=\ensuremath{\mathcal{N}},
    description={A set of agents},
    sort={Set of agents},
    type=symbol
}
\newcommand{\setOfAgents}{\gls{rl:setOfAgents}}

\newglossaryentry{rl:actionSpace}{
    name=\ensuremath{\mathcal{A}},
    description={Action space},
    sort={Action space},
    type=symbol
}
\newcommand{\actionSpace}{\gls{rl:actionSpace}}

\newglossaryentry{rl:stateSpace}{
    name=\ensuremath{\mathcal{S}},
    description={State space},
    sort={State space},
    type=symbol
}
\newcommand{\stateSpace}{\gls{rl:stateSpace}}

\newglossaryentry{rl:observationSpace}{
    name=\ensuremath{\mathcal{O}},
    description={Observation space},
    sort={Observation space},
    type=symbol
}
\newcommand{\observationSpace}{\gls{rl:observationSpace}}

\newglossaryentry{rl:observationFcn}{
    name=\ensuremath{\Omega},
    description={Observation function},
    sort={Observation function},
    type=symbol
}

\newglossaryentry{rl:transitionProbFcn}{
    name=\ensuremath{\mathcal{P}},
    description={Transition probability function},
    sort={Transition probability function},
    type=symbol
}
\newcommand{\transitionProbFcn}{\gls{rl:transitionProbFcn}}

\newglossaryentry{rl:rewardFcn}{
    name=\ensuremath{R},
    description={Reward function},
    sort={Reward function},
    type=symbol
}
\newcommand{\rewardFcn}{\gls{rl:rewardFcn}}

\newglossaryentry{rl:discountFactor}{
    name=\ensuremath{\gamma},
    description={Discount factor},
    sort={Discount factor},
    type=symbol
}
\newcommand{\discountFactor}{\gls{rl:discountFactor}}

\newglossaryentry{rl:state}{
    name=\ensuremath{s},
    description={State},
    sort={State},
    type=symbol
}
\newcommand{\state}{\gls{rl:state}}

\newglossaryentry{rl:nextState}{
    name=\ensuremath{s'},
    description={Next state},
    sort={Next state},
    type=symbol
}

\newglossaryentry{rl:action}{
    name=\ensuremath{a},
    description={Action},
    sort={Action},
    type=symbol
}
\newcommand{\action}{\gls{rl:action}}

\newglossaryentry{rl:jointActions}{
    name=\ensuremath{\bm{a}},
    description={Joint actions},
    sort={Joint actions},
    type=symbol
}
\newcommand{\jointActions}{\gls{rl:jointActions}}

\newglossaryentry{rl:observation}{
    name=\ensuremath{o},
    description={Observation},
    sort={Observation},
    type=symbol
}
\newcommand{\observation}{\gls{rl:observation}}

\newglossaryentry{rl:jointObservations}{
    name=\ensuremath{\bm{o}},
    description={Joint observations},
    sort={Joint observations},
    type=symbol
}
\newcommand{\jointObservations}{\gls{rl:jointObservations}}

\newglossaryentry{rl:reward}{
    name=\ensuremath{r},
    description={Reward},
    sort={Reward},
    type=symbol
}

\newglossaryentry{rl:jointRewards}{
    name=\ensuremath{\bm{r}},
    description={Joint rewards},
    sort={Joint rewards},
    type=symbol
}
\newcommand{\jointRewards}{\gls{rl:jointRewards}}

\newglossaryentry{rl:valueFunction}{
    name=\ensuremath{V},
    description={A function that evaluates how good a policy is},
    sort={Value function},
    type=symbol
}
\newcommand{\valueFunction}{\gls{rl:valueFunction}}

\newglossaryentry{rl:policy}{
    name=\ensuremath{\pi},
    description={Policy},
    sort={Policy},
    type=symbol
}
\newcommand{\policy}{\glslink{rl:policy}{\ensuremath{\pi}}}

\newglossaryentry{sym:distance}{
    name=\ensuremath{d},
    description={Distance},
    sort={Distance},
    type=symbol
}
\newcommand{\distance}{\gls{sym:distance}}

\newglossaryentry{sym:numOfPointsRef}{
    name=\ensuremath{n_\text{p,RP}},
    description={Number of points on the reference path},
    sort={Number of points on the reference path},
    type=symbol
}
\newcommand{\numOfPointsRef}{\gls{sym:numOfPointsRef}}

\newglossaryentry{sym:numOfObservedSurroundingAgents}{
    name=\ensuremath{n_\text{sur.}},
    description={Number of observed surrounding agents},
    sort={Number of observed surrounding agents},
    type=symbol
}
\newcommand{\numOfObservedSurroundingAgents}{\gls{sym:numOfObservedSurroundingAgents}}

\newglossaryentry{sym:numOfNotMaskedSurroundingAgents}{
    name=\ensuremath{n_\text{sur.,NM}},
    description={Number of not masked surrounding agents},
    sort={Number of not masked agents},
    type=symbol
}

\newglossaryentry{sym:numOfMaskedSurroundingAgents}{
    name=\ensuremath{n_\text{sur.,M}},
    description={Number of masked surrounding agents},
    sort={Number of masked agents},
    type=symbol
}

\newglossaryentry{rl:numOfSamples}{
    name=\ensuremath{n_\text{samples}},
    description={Number of training samples},
    sort={Number of training samples},
    type=symbol
}

\newglossaryentry{rl:setOfModels}{
    name=\ensuremath{\mathcal{M}},
    description={A set of models},
    sort={A set of models},
    type=symbol
}
\newcommand{\setOfModels}{\gls{rl:setOfModels}}

\newglossaryentry{rl:maxModelIndex}{
    name=\ensuremath{5},
    description={Maximum Model Index},
    sort={Maximum Model Index},
    type=symbol
}
\newcommand{\maxModelIndex}{\gls{rl:maxModelIndex}}

\newglossaryentry{rl:numberOfModels}{
    name=\text{six},
    description={Number of models},
    sort={Number of models},
    type=symbol
}
\newcommand{\numberOfModels}{\gls{rl:numberOfModels}}

\newglossaryentry{rl:model}{
    name=\ensuremath{M},
    description={RL Model},
    sort={RL Model},
    type=symbol
}
\newcommand{\model}{\gls{rl:model}}

\newglossaryentry{rl:compositeScore}{
    name=\ensuremath{CS},
    description={Composite Score},
    sort={Composite Score},
    type=symbol
}
\newcommand{\compositeScore}{\gls{rl:compositeScore}}

\newglossaryentry{rl:collisionRate}{
    name=\ensuremath{CR},
    description={Collision Rate},
    sort={Collision Rate},
    type=symbol
}
\newcommand{\collisionRate}{\gls{rl:collisionRate}}

\newglossaryentry{rl:collisionRateTotal}{
    name=\ensuremath{CR_{\text{total}}},
    description={Total Collision Rate},
    sort={Total Collision Rate},
    type=symbol
}
\newcommand{\collisionRateTotal}{\gls{rl:collisionRateTotal}}

\newglossaryentry{rl:collisionRateAA}{
    name=\ensuremath{CR_{\text{A-A}}},
    description={Agent-Agent Collision Rate},
    sort={Agent-Agent Collision Rate},
    type=symbol
}
\newcommand{\collisionRateAA}{\gls{rl:collisionRateAA}}

\newglossaryentry{rl:collisionRateAL}{
    name=\ensuremath{CR_{\text{A-L}}},
    description={Agent-Lanelet Collision Rate},
    sort={Agent-Lanelet Collision Rate},
    type=symbol
}
\newcommand{\collisionRateAL}{\gls{rl:collisionRateAL}}

\newglossaryentry{rl:centerLineDeviation}{
    name=\ensuremath{CD},
    description={Center Line Deviation},
    sort={Score: Center Line Deviation},
    type=symbol
}
\newcommand{\centerLineDeviation}{\gls{rl:centerLineDeviation}}

\newglossaryentry{rl:averageSpeed}{
    name=\ensuremath{AS},
    description={Average Speed},
    sort={Average Speed},
    type=symbol
}
\newcommand{\averageSpeed}{\gls{rl:averageSpeed}}

\DeclareAcronym{cav}{
    short = CAV,
    long  = Connected and Automated Vehicle,
}
\DeclareAcronym{cg}{
    short = CG,
    long = Center of Gravity,
    short-plural = s,
    long-plural-form = Centers of Gravity,
}
\DeclareAcronym{cnn}{
    short = CNN,
    long  = Convolutional Neural Network
}
\DeclareAcronym{cpm}{
    short = CPM,
    long  = Cyber-Physical Mobility
}
\DeclareAcronym{cpmlab}{
    short = CPM Lab,
    long  = Cyber-Physical Mobility Lab
}
\DeclareAcronym{dmpc}{
    short = DMPC,
    long  = distributed model predictive control
}
\DeclareAcronym{dql}{
    short = DQL,
    long  = Deep Q-Learning
}

\DeclareAcronym{il}{
    short = IL,
    long  = Imitation Learning,
    short-indefinite = an,
}
\DeclareAcronym{mas}{
    short = MAS,
    long  = Multi-Agent System,
    short-indefinite = an,
}
\DeclareAcronym{mdp}{
    short = MDP,
    long  = Markov Decision Process,
    short-indefinite = an,
}
\DeclareAcronym{mg}{
    short = MG,
    long  = Markov Game,
    short-indefinite = an,
}
\DeclareAcronym{ml}{
    short = ML,
    long  = Machine Learning,
    short-indefinite = an,
}
\DeclareAcronym{mpc}{
    short = MPC,
    long  = model predictive control,
    short-indefinite = an,
}
\DeclareAcronym{marl}{
    short = MARL,
    long  = Multi-Agent Reinforcement Learning,
    short-indefinite = an,
}
\DeclareAcronym{ocp}{
    short = OCP,
    long  = optimal control problem,
    short-indefinite = an,
    long-indefinite = an,
}
\DeclareAcronym{per}{
    short = PER,
    long  = Prioritized Experience Replay
}
\DeclareAcronym{pdmpc}{
    short = \mbox{P-DMPC},
    long  = prioritized \acl{dmpc}
}
\DeclareAcronym{pomdp}{
    short = POMDP,
    long  = Partially Observable \ac{mdp}
}
\DeclareAcronym{pomg}{
    short = POMG,
    long  = Partially Observable Markov Game
}
\DeclareAcronym{ppo}{
    short = PPO,
    long  = Proximal Policy Optimization
}
\DeclareAcronym{rhc}{
    short = RHC,
    long  = receding horizon control,
    short-indefinite = an,
}
\DeclareAcronym{rl}{
    short = RL,
    long  = Reinforcement Learning,
    short-indefinite = a,
}
\DeclareAcronym{zsg}{
    short = ZSG,
    long  = Zero-Shot Generalization,
}
\newglossaryentry{def:agent}{
	name=agent,
	description={An agent is a system which is composed of at least one of the three elements: sensors, actuators, and a dynamic behavior.%
    },
}

\newglossaryentry{def:agentActive}{
	name=active agent,
	description={Active \glspl{def:agent} are \glspl{def:agent} which are connected using a communication
    network over which they can exchange data. The exchanged data is
    used by the \glspl{def:agent}' controllers to find appropriate inputs to achieve their
    goals while interacting with other \glspl{def:agent}.
    Additionally, active \glspl{def:agent} consider \glspl{def:agentPassive}},
    parent=def:agent,
}

\newglossaryentry{def:agentPassive}{
	name=passive agent,
	description={Passive \glspl{def:agent} are \glspl{def:agent} without networked control. However, they can communicate their data like current and future states to \glspl{def:agentActive}, or they can be detected by \glspl{def:agentActive}' sensors.%
    },
    parent=def:agent,
}

\newglossaryentry{def:distrutedSolutionQuality}{
	name=distributed solution quality,
	description={%
        The quality $q\in [0,1]$ of the solution in \ac{dmpc} is the networked objective function value ${\fcnObjective}_{c}$ for the solution of the corresponding \ac{cmpc} formulation divided by the objective function value ${\fcnObjective_d}$ for the solution of the \ac{dmpc} formulation
        \begin{equation}
            q = \frac{\fcnObjective_c}{\fcnObjective_d}.
        \end{equation}
    },
}

\newglossaryentry{def:mas}{
	name=multi-agent system,
	description={A system consisting of multiple \glspl{def:agent}.%
    },
}

\newglossaryentry{def:ncs}{
	name=networked control system,
	description={A system consisting of multiple \glspl{def:agentActive}.%
    },
}

\newglossaryentry{def:prediction}{
	name=prediction,
	description={
        A prediction $\agentPrediction^{\anAgent}_{\cdot\vert \timestep}$ of \gls{def:agent} $\anAgent$ is its predicted state trajectory as obtained from the solution of its \ac{ocp} at time $\timestep$.
        A prediction $\agentPredictionForAgentAFromAgentBAtTimeC{\anAgent}{\anotherAgent}{\timestep}$ of \gls{def:agent} $\anAgent$ for \gls{def:agent} $\anotherAgent$ is agent $\anotherAgent$'s state trajectory as viewed from agent $\anAgent$ at time $\timestep$. It is obtained by communication or by predicting \gls{def:agent} $\anotherAgent$'s state trajectory with its model using the solution to its \ac{ocp}.%
    },
}

\newglossaryentry{def:consistency}{
	name=prediction consistency,
	description={%
        \Iac{ncs} is prediction consistent at time step $\timestep$ if the \gls{def:prediction} \agentPredictionForAgentAFromAgentBAtTimeC{\anotherAgent}{\anAgent}{\timestep} of every agent $\anAgent\in\setAgents$ for each of its neighbors $\anotherAgent \in \setNeighbors{\anAgent}$ equals the actual \gls{def:prediction} $\agentPrediction^{(\anotherAgent)}_{\cdot\vert \timestep}$ of its neighbors, i.e.,
        \begin{equation}
            \agentPredictionForAgentAFromAgentBAtTimeC{\anotherAgent}{\anAgent}{\timestep}=\agentPrediction^{(\anotherAgent)}_{\cdot\vert \timestep}, \quad \forall \anAgent \in \setAgents, \forall \anotherAgent \in \setNeighbors{\anAgent}.
        \end{equation}%
    }
}

\newglossaryentry{def:ncsFeasible}{
	name=NCS-feasible,
	description={
        The solutions $\sysControlInputs_{\cdot \vert \timestep}\ofAgent{\anAgent}$ of all agents $i\in\setAgents$ are \acs*{ncs}-feasible if the stacked solution vector $\bm{U}_{\cdot \vert \timestep} = \left( \sysControlInputs_{\cdot \vert \timestep}\ofAgent{1}, \ldots, \sysControlInputs_{\cdot \vert \timestep}\ofAgent{\numAgents} \right)\transposed$ satisfies all constraints of the corresponding central \acf*{ocp} considering all agents.%
    },
}

\newglossaryentry{def:feasibleAgent}{
	name=agent-feasible,
	description={%
        A solution is agent-feasible if it satisfies the constraints of to the corresponding agent's \ac{ocp}.%
    },
}

\newglossaryentry{def:networkedObjectiveFunction}{
	name=networked objective function,
	description={%
        The objective function value ${\fcnObjective}$ in \iac{ncs} formulation is the sum of all agent objective functions \fcnObjective[i]
        \begin{equation}
            \fcnObjective = \sum_{i}^{i\in\setAgents} \fcnObjective[i].
        \end{equation}
    },
}

\newglossaryentry{def:optimalPriorityAssignment}{
	name=optimal priority assignment,
	description={%
        The optimal priority assignment results in a feasible solution for every agent with the lowest networked objective function value.%
    },
}

\newglossaryentry{def:graph}{
	name=graph,
	description={%
        A directed graph $\graphDirected = \left(\setVertices,\setEdges\right)$ is a pair of two sets,
        the set of vertices $\setVertices=\set{1,\dots,\numAgents}$
        and the set of directed edges $\setEdges \subseteq \setVertices \times \setVertices$.
        The edge from $i$ to $j$ is denoted by $\edgeDirected{i}{j}$.
        An undirected graph $\graphUndirected = \left(\setVertices,\setEdges\right)$ is a special form of a directed graph in which every edge is directed both ways, i.e., $\edgeDirected{i}{j} \in \setEdges \iff \edgeDirected{j}{i} \in \setEdges$.
    },
}

\newglossaryentry{def:path}{
	name=path,
	description={%
        A path of a graph $\graphDirected$ is a subgraph $\graphDirected_{\graphPath} = \left(\setVertices_{\graphPath},\setEdges_{\graphPath}\right)\subseteq\graphDirected$ with distinct vertices $\setVertices_{\graphPath}=\{i_{1},i_{2},i_{3},\ldots,i_{\numVerticesPath-1},i_{\numVerticesPath}\}$ and distinct edges $\setEdges_{\graphPath}=\{\edgeDirected{i_{1}}{i_{2}},\edgeDirected{i_{2}}{i_{3}},\ldots,\edgeDirected{i_{\numVerticesPath-1}}{i_{\numVerticesPath}}\}$, with $\numVerticesPath$ being the number of vertices of the path. The length of the path is defined as $\numVerticesPath-1$.
    },
}

\newglossaryentry{def:graph:adjacency}{
	name=adjacency,
	description={%
    A vertex $j$ is a predecessor of vertex $i$ iff $\edgeDirected{j}{i}\in\setEdges$.
    The set of predecessors of vertex $i$ is denoted by
    \begin{equation}
        \setPredecessors{i}=\set{j \mid \edgeDirected{j}{i}\in\setEdges}.
    \end{equation}
    A vertex $j$ is a successor of vertex $i$ iff $\edgeDirected{i}{j}\in\setEdges$.
    The set of successors of vertex $i$ is denoted by
    \begin{equation}
        \setSuccessors{i}=\set{j \mid \edgeDirected{i}{j}\in\setEdges}.
    \end{equation}
    A vertex $j$ is a neighbor to or adjacent to vertex $i$ if it is either a predecessor or a successor.
    The set of neighbors of vertex $i$ is denoted by
    \begin{equation}
        \setNeighbors{i}= \setSuccessors{i} \cup \setPredecessors{i}.
    \end{equation}
    },
    parent=def:graph,
}

\newglossaryentry{def:graph:degree}{
	name=degree,
	description={%
        The degree $\vertexDegree{i} = \lvert \setNeighbors{i} \rvert$ denotes the number of the adjacent vertices of vertex $i$. 
        The number of incoming edges called in-degree is denoted by $\vertexInDegree{i} = \lvert \setPredecessors{i} \rvert$.
        The number of outgoing edges called out-degree is denoted by $\vertexOutDegree{i} = \lvert \setSuccessors{i} \rvert$.%
    },
    parent=def:graph,
}

\newglossaryentry{def:couplingGraph}{
	name=coupling graph,
	description={A coupling graph $\graphDirected=(\setVertices,\setEdges)$ is a graph that represents the interaction between agents. Vertices represent agents and edges denote coupling objectives or constraints. A vertex from agent $i$ to agent $j$ corresponds to a coupling objective or constraint in the \ac{ocp} of agent $j$.%
    },
}

\newglossaryentry{def:matrix:Adjacency}{
	name=adjacency matrix,
	description={An adjacency matrix represents a graph with $\numAgents$ vertices in a matrix $\matAdjacency \in \set{0,1}^{\numAgents\times\numAgents}$ with entries
    \begin{equation}
        \matAdjacencyElement{ij} =
            \begin{cases}
                1 & \text{ if } \edgeDirected{i}{j} \in \setEdges \\
                0 & \text{ otherwise.}
            \end{cases}
    \end{equation}
    },
}

\newglossaryentry{def:tCompNcs}{
	name=computation time of \iac{ncs},
	description={%
        The computation time $\tCompNcs$ of \iac{ncs} is the time required for the \ac{ncs} to measure the states, formulate and solve the \ac{ocp}, apply the inputs to all agents, and communicate the required data. 
    },
}

\newglossaryentry{def:setReachable}{
	name=reachable set,
	description={%
        The reachable set of states $\setReachable$ of an agent from an initial time $t_{\text{init.}}$ to an end time $t_{\text{end}}$ is
            \begin{equation}\label{eq:setReachable}
                \setReachable_{[t_{\text{init.}},t_{\text{end}}] \mid t_{\text{init.}}} = \biggl\{ \int_{t_{\text{init.}}}^{t_{\text{end}}} \sysModelContinuous(\sysState,\sysControlInputs)dt
                \biggm| \sysState(t_{\text{init.}}) \in \setFeasibleStates(t_{\text{init.}}), \forall t: \sysControlInputs \in \setFeasibleInputs \biggr\},
            \end{equation}
        with the possible system initial states $\sysState(t_{\text{init.}})$ being bounded by its initially admissible set $\setFeasibleStates(t_{\text{init.}}) \subseteq \setRealNumbers^{\numStates}$, and the possible system control inputs $\sysControlInputs$ being bounded by its admissible set $\setFeasibleInputs \subseteq \setRealNumbers^{\numInputs}$.
    },
}

\newglossaryentry{def:conflictualDecisions}{
	name=conflictual decisions in \iac{ncs},
	description={%
        Consider two decisions made by two agents of \iac{ncs} at time step $\timestep$ with a duration $N_k$.
        They are deemed conflictual if the predicted outcome of the decisions violates the \ac{ncs}-feasibility at some point in time.%
    },
}

\newglossaryentry{def:conflictualSpace}{
	name=conflictual space of \iac{ncs},
	description={%
        In dynamic systems, the state space represents the set of all possible states the systems can occupy. 
        The conflictual space refers to a subset, or potentially the entirety, of this state space where whether decisions are conflictual is determined.
    },
}

\input{copyright.tex}

\def\BibTeX{{\rm B\kern-.05em{\sc i\kern-.025em b}\kern-.08em
  T\kern-.1667em\lower.7ex\hbox{E}\kern-.125emX}}

\begin{document}
\title{\LARGE \bf
    SigmaRL: A Sample-Efficient and Generalizable Multi-Agent Reinforcement Learning Framework for Motion Planning
    \thanks{This research was supported by the Bundesministerium für Digitales und Verkehr (German Federal Ministry for Digital and Transport) within the project ``Harmonizing Mobility'' (grant number 19FS2035A).}
}

\author{
    Jianye Xu$^{1}$\,\orcidlink{0009-0001-0150-2147},~\IEEEmembership{Student~Member,~IEEE},
    Pan Hu$^{2}$\,\orcidlink{0009-0004-6683-8778},
    Bassam Alrifaee$^{3}$\,\orcidlink{0000-0002-5982-021X},~\IEEEmembership{Senior Member, ~IEEE}
    \thanks{$^{1}$The author is with the Chair of Embedded Software (Informatik 11), RWTH Aachen University, Germany, \href{mailto:xu@embedded.rwth-aachen.de}{\tt\footnotesize xu@embedded.rwth-aachen.de}.}
    \thanks{$^{2}$The author is with the Department of Computer Science, RWTH Aachen University, Germany, \href{mailto:pan.hu@rwth-aachen.de}{\tt\footnotesize pan.hu@rwth-aachen.de}.}
    \thanks{$^{3}$The author is with the Department of Aerospace Engineering, University of the Bundeswehr Munich, Germany, \href{mailto:bassam.alrifaee@unibw.de}{\tt\footnotesize bassam.alrifaee@unibw.de}.}
}

    \maketitle
\thispagestyle{IEEEtitlepagestyle}
\begin{abstract}
This paper introduces an open-source, decentralized framework named {\it SigmaRL}, designed to enhance both \underline{s}ample eff\underline{i}ciency and \underline{g}eneralization of \underline{m}ulti-\underline{a}gent \underline{R}einforcement \underline{L}earning (RL) for motion planning of connected and automated vehicles.
Most RL agents exhibit a limited capacity to generalize, often focusing narrowly on specific scenarios, and are usually evaluated in similar or even the same scenarios seen during training.
Various methods have been proposed to address these challenges, including experience replay and regularization.
However, how observation design in RL affects sample efficiency and generalization remains an under-explored area.
We address this gap by proposing five strategies to design information-dense observations, focusing on general features that are applicable to most traffic scenarios.
We train our RL agents using these strategies on an intersection and evaluate their generalization through numerical experiments across completely unseen traffic scenarios, including a new intersection, an on-ramp, and a roundabout.
Incorporating these information-dense observations reduces training times to under one hour on a single CPU, and the evaluation results reveal that our RL agents can effectively zero-shot generalize.
\par\medskip
\noindent
Code: \href{https://github.com/bassamlab/SigmaRL}{\small github.com/bassamlab/SigmaRL}
\end{abstract}

\acresetall

\section{Introduction}
\subsection{Motivation}\label{sec:motivation}
\Ac{rl} has become an increasingly promising approach for motion planning of \acp{cav}, owing to its ability to learn through interactions with the environment.
Despite its great success, the generalization of RL agents---i.e., the ability to generalize to unseen scenarios or environments---remains one of the fundamental challenges.

Most RL agents for \acp{cav} are specialized for a specific scenario, such as an intersection, an on-ramp, or lane-following, see~\cite{aradi2022survey} for a comprehensive survey.
Besides, due to a lack of generalization, they are often tested in a similar or even the same scenario seen during training.
This leads to an RL agent being only used in one scenario.
It may even fail to generalize to new scenarios or tasks that seem similar to the training scenarios or tasks~\cite{farebrother2020generalization}.

To enhance their generalization, one strategy involves enriching the diversity of training scenarios, e.g., training in an environment where different scenarios are involved.
However, this strategy tends to impair sample efficiency, as most traffic situations are not challenging, owing to the rarity of safety-critical events~\cite{feng2023dense}.
We call this inefficiency \textit{sample inefficient}.
A \textit{sample} (also called an experience or a frame) in \ac{rl} refers to a single interaction instance between an agent and an environment. 
This interaction includes the agent observing the state of the environment, taking an action based on its policy, and receiving feedback in the form of a reward and a new state from the environment.
The learning agent uses these samples to update its policy.
By \textit{inefficient}, we mean that the learning agent requires an excessive amount of samples before the policy achieves a satisfactory performance.
Formally, we define sample efficiency as follows.
\begin{definition}[Sample Efficiency]\label{def:sampleEfficiency}
    Consider a set of RL models $\mathcal{M}$, each trained with a fixed number of samples.
    For each model $\model_i \in \setOfModels$, we use the performance metrics:
    \begin{itemize}
        \item \textbf{Collision Rate} ($\collisionRate_{\model_i}$): The proportion of time steps where agents cause a collision,
        \item \textbf{Center Line Deviation} ($\centerLineDeviation_{\model_i}$): The average deviation of all agents from their lane center lines, and
        \item \textbf{Average Speed} ($\averageSpeed_{\model_i}$): The average speed of all agents
    \end{itemize}
    to quantify its sample efficiency by a composite score
    \begin{equation}\label{eq:score}
        \compositeScore_{\model_i} \coloneq -w_1 \cdot \collisionRate_{\model_i} - w_2 \cdot \centerLineDeviation_{\model_i} + w_3 \cdot \averageSpeed_{\model_i},
    \end{equation}
    where $ w_1$, $w_2$, and $w_3$ are weighting factors to balance the relative importance and scale of the performance metrics.
    The model $\model^* = \arg\max_{\model_i \in \mathcal{M}} \compositeScore_{\model_i}$ that maximizes the composite score $\compositeScore_{\model_i}$ among all models $ \model_i \in \setOfModels $ is deemed the most sample-efficient.
\end{definition}

The above-mentioned sample inefficiency is further amplified by the reliance on raw-sensor-data-based end-to-end learning paradigms, requiring RL agents to learn a direct mapping from raw sensor data to actions.
Direct learning from raw sensor data may allow agents to uncover patterns and correlations that might be missed with handcrafted features.
However, it demands sophisticated feature extraction capabilities, typically provided by deep \acp{cnn}, which significantly increases the complexity of the learning process.

In light of these challenges, we identify a need to develop an effective approach toward sample-efficient and generalizable RL for \acp{cav}.

\subsection{Related Work}\label{sec:relatedWork}
Despite the success of RL, the training process often demands an extensive amount of samples in real-world applications~\cite{duan2016rl2}, ranging from tens to hundreds of millions, casting sample efficiency important ongoing research.
Additionally, generalization is vital for RL agents, especially in real-world applications where environments are unpredictable and it is impractical to generate training data that can cover all possible situations.
Consequently, generalization in RL has gained substantial attention in recent years.
Below, we briefly overview recent works that address sample efficiency and generalization in RL.

\textbf{Sample Efficiency:}
In~\cite{yu2018sample}, the sample efficiency of RL from multiple aspects is discussed, such as more effective environmental exploration and better policy optimization.
Another concern in RL is the sequential dependency of experiences---an alternative term for samples, which violates the independent and identically distributed (i.i.d.) assumption that underlies many stochastic gradient descent algorithms for RL. 
Experience replay in~\cite{lin1992selfimproving} addresses this concern by storing experiences in a memory buffer and randomly sampling from it when learning, instead of learning directly from incoming experiences.
This method largely enhances sample efficiency, making it a new standard in many RL algorithms~\cite{zhang2018deeper}. 
To make learning more efficient,~\cite{schaul2016prioritized} proposes a strategy called \ac{per}, which assigns greater importance to certain experiences based on a prioritization scheme, ensuring that more crucial experiences are sampled with higher frequency.
Two recent works~\cite{kaufmann2023championlevel,song2023reaching} apply RL for autonomous drone racing, showcasing champion-level racing performance in their experiments.
It uses an initial state buffer to store successful states for agents navigating through a gate, and it has been highlighted that sampling from this buffer during environment resets significantly enhances sample efficiency.
Another work~\cite{buckman2018sampleefficient} proposes an approach for integrating model-free and model-based RL to combine the high performance of the former with the reduced sample complexity of the latter. Further,~\cite{du2020good} examines the sufficiency of a good representation of function approximation for achieving sample-efficient RL, particularly regarding value functions, transition models, reward functions, and policies.
Despite these advances, a gap remains in research specifically targeting the design of observations to boost the sample efficiency of RL agents.

\textbf{Generalization}: 
The authors in~\cite{farebrother2020generalization} explore the potential of regularization techniques such as dropout to enhance both the sample efficiency and generalization of \ac{dql} and show that it can foster the learning of more general features.
Meanwhile,~\cite{packer2019assessing} presents an empirical comparison of two prominent RL algorithms, A2C~\cite{mnih2016asynchronous} and \ac{ppo}~\cite{schulman2017proximal}, examining their generalization in conjunction with EPOpt~\cite{rajeswaran2017epopt} and RL$^2$~\cite{duan2016rl2}, two methods aimed at addressing the generalization problem of RL.
Further,~\cite{cobbe2019quantifying} introduces an RL environment called CoinRun, a benchmark for assessing the generalization of RL. This benchmark provides metrics to quantify how various factors---like neural network structures and regularization techniques---impact the generalization.
In~\cite{li2023metadrive}, a simulation platform called MetaDrive is developed to facilitate the research of generalizable RL agents for autonomous driving, accommodating both single- and multi-agent settings. In addition, the conducted experiments therein show that diversifying training scenarios improves the generalization of RL agents.
Observing the non-stationarity in on-policy RL,~\cite{igl2020transient} proposes a hypothesis that neural networks exhibit a memory effect, which can harm the generalization when training data distributions shift. It proposes an approach called Iterate Relearning to counteract this effect.
Zero-shot generalization represents a specific type of generalization, referring to an RL agent's ability to generalize to completely unseen scenarios without further training or fine-tuning.
A recent survey~\cite{kirk2023survey} categorizes approaches to realize zero-shot generalization, highlighting strategies to either increase the similarity, decrease the difference between training and testing data, or enhance optimization to prevent overfitting.
Nevertheless, the majority of the state-of-the-art research focuses on improving the generalization of RL agents on the algorithm level.

In summary, while substantial research efforts have contributed to enhancing RL agents' sample efficiency and generalization, a notable research gap remains in exploring how observation design could further improve them.
Observation design in RL, particularly for \acp{cav}, is crucial because it determines the quality of environmental information that the learning agents receive.
Poorly designed observations can lead to inefficient and ineffective learning, as the agents might incorrectly interpret their surroundings.

\subsection{Paper Contributions}\label{sec:contribution}
The main contributions of this paper are twofold:
\begin{itemize}
    \item It presents an open-source, decentralized \ac{marl} framework named \texttt{SigmaRL} for motion planning of \acp{cav}. The RL agents within this framework require less than one hour of training time on a single CPU and can zero-shot generalize to completely unseen traffic scenarios.
    \item It examines how observation design influences sample efficiency and generalization of RL agents---an under-explored area within the RL community. As outcomes, it proposes five strategies for designing information-dense, structured observations for motion planning, opposite to bulky, image-based observations. They are
    \begin{enumerate*}
        \item using an ego view instead of a bird-eye view, 
        \item observing vertices of surrounding agents instead of their poses and geometric dimensions,
        \item observing distances to surrounding agents,
        \item observing distances to lane boundaries instead of points sampled from them, and
        \item observing distances to lane center lines.
    \end{enumerate*}
\end{itemize}

\subsection{Notation}\label{sec:notation}
A variable $x$ is marked with a superscript $x^{(\anAgent)}$ if it belongs to agent $\anAgent$.
All other information is presented in its subscript. For example, the value of $x$ at time $\timeStep$ is written as $x\ofATimeStep$.
If multiple pieces of information need to be conveyed in subscript, they are separated by commas.
For any set $\mathcal{S}$, the cardinality of the set is denoted by $|\mathcal{S}|$.
We use the term \texttt{agent} to refer interchangeably to a vehicle or an RL agent.

\subsection{Paper Structure}\label{sec:structure}
\Cref{sec:problemFormulation} formally formulates the problem.
\Cref{sec:framework} presents our framework as a solution, including the RL algorithm, the environment, the observation design, and the reward design.
\Cref{sec:ablationStudies} details experiment setups and discusses the experiment results.
\Cref{sec:conclusions} draws conclusions and outlines future research.

\section{Problem Formulation}\label{sec:problemFormulation}
We consider the problem of \ac{marl} for motion planning of \acp{cav} in a discrete-time setting, where a set of agents interact in a shared environment, aiming to achieve both safety and efficiency in traffic flow.
This problem can be described using a Markov Game, also known as a stochastic game~\cite{shapley1953stochastic}, which is a standard multi-agent setting of a Markov Decision Process.
In addition, we consider agents have limited sensor capabilities, and they can only sense their surrounding environments, making the problem more realistic.
This condition is known as partial observability, leading to the so-called \ac{pomg}, which is challenging to solve due to imperfect information of the game~\cite{oliehoek2016concise}.
Formally, we define a \ac{pomg} as follows.
\begin{definition}[Adapted from~\cite{hansen2024dynamic}]\label{pomg}
    A \ac{pomg} is defined by a tuple 
    $(
        \setOfAgents, 
        \stateSpace, 
        \{\actionSpace\ofAnAgent\}_{\anAgent \in \setOfAgents},
        \{\observationSpace\ofAnAgent\}_{\anAgent \in \setOfAgents},
        \transitionProbFcn,  
        \{\rewardFcn\ofAnAgent\}_{\anAgent \in \setOfAgents}, 
        \discountFactor
    )$, where
    \begin{itemize}
        \item $\setOfAgents = \{1, \dots, N\}$ denotes a finite set of agents.
        \item $\stateSpace$ is the state space of the system shared by all agents.
        \item $\actionSpace\ofAnAgent$ denotes the action space of agent $\anAgent$. The joint action space of all agents is the Cartesian product: $\actionSpace \coloneq \times_{\anAgent \in \setOfAgents} \actionSpace\ofAnAgent$. The joint action of all agents at time step $\timeStep$ is denoted as 
        $\jointActions\ofATimeStep \coloneq (\action\ofATimeStep\ofAgent{1},\ldots,\action\ofATimeStep\ofAgent{N})$, 
        where $\action\ofATimeStep\ofAnAgent$ denotes the action of agent $\anAgent$ at time step $\timeStep$.
        \item $\observationSpace\ofAnAgent$ is the observation space of agent $\anAgent$. 
        Similarly, $\observationSpace \coloneq \times_{\anAgent \in \setOfAgents} \observationSpace\ofAnAgent$ and $\jointObservations\ofATimeStep$ denote the joint observation space and the joint observations of all agents at time step $\timeStep$, respectively.
        \item $\transitionProbFcn: \stateSpace \times \actionSpace \rightarrow \Delta(\stateSpace) \times \Delta(\observationSpace)$ is the Markovian state transition and observation probability function, which describes the probability of transitioning from one state to another and agents getting certain observations given the current state and the joint actions of all agents. $\Delta$ denotes a probability distribution over a space.
        \item $\rewardFcn\ofAnAgent: \stateSpace \times \actionSpace \times \stateSpace \rightarrow \setRealNumbers$ is the reward function that determines the immediate reward received by agent $\anAgent$ for a transition from $(\state\ofATimeStep, \jointActions\ofATimeStep)$ to $\state\ofNextTimeStep$, where $\state\ofATimeStep$ and $\state\ofNextTimeStep$ are the environment states before and after the transition, and $\jointActions\ofATimeStep$ is the joint action. We denote the joint rewards at time step $\timeStep$ with $\jointRewards\ofATimeStep$.
        \item $\gamma \in [0,1)$ is the discount factor that balances the immediate and future rewards.
    \end{itemize}
\end{definition}

At any given time step $\timeStep$, each agent $\anAgent$ undertakes an action $\action\ofAnAgent\ofATimeStep$ in response to its partial observation $\observation\ofAnAgent\ofATimeStep$.
Following this, the environment transitions to a new state $\state\ofNextTimeStep$ and rewards each agent $\anAgent$ through the reward function $\rewardFcn\ofAnAgent(\state\ofATimeStep, \action\ofAnAgent\ofATimeStep, \state\ofNextTimeStep)$.

\begin{problem}\label{prob:generalizableAgents}
    Considering a \ac{pomg} defined in \cref{pomg}, design partial observations that allow each agent $\anAgent$ to efficiently learn a policy $\policy\ofAnAgent: O_i \rightarrow \Delta(\actionSpace\ofAnAgent)$, which is a mapping from its partial observation $\observation\ofAnAgent$ to a probability distribution over its action space $\actionSpace\ofAnAgent$.
    The design of partial observations should maximize the sample efficiency defined in \cref{def:sampleEfficiency}.
    The agents should be applicable to motion planning of \acp{cav} and can be zero-shot generalized to unseen traffic scenarios.
\end{problem}

Note that the single-agent setting of a \ac{pomg} is already Nondeterministic Exponential Time Complete (NEXP-Complete), requiring super-exponential time to find the optimal solution in the worst case~\cite{bernstein2002complexity}.
Instead of trying to find the optimal solution, we focus on how to design observations to improve the sample efficiency and generalization of RL agents.

\section{Our Framework}\label{sec:framework}
In this section, we detail our framework \texttt{SigmaRL}.
We describe its RL algorithm in \cref{sec:rlAlgorithm} and its RL environment in \cref{sec:rlEnv}.
We propose the observation-design strategies in \cref{sec:observationDesign}.
Reward design is out of our scope in this work, and we refer to our open-source repository for details.

\subsection{RL Algorithm}\label{sec:rlAlgorithm}
We employ a multi-agent extension of \ac{ppo}, termed multi-agent \ac{ppo}~\cite{lowe2017multiagent}, as our RL algorithm.
PPO is a widely adopted RL algorithm that uses gradients to adjust policies toward more rewarding actions, featuring an architecture that includes both an actor to make decisions and a critic to evaluate these decisions.
Multi-agent \ac{ppo} extends it to a multi-agent setting, using a centralized critic alongside distributed actors, a popular learning scheme known as centralized-learning-decentralized-execution.
Note that the centralized critic is only needed during training.
This scheme is particularly designed to mitigate the non-stationary problem in \ac{marl}, especially under conditions of partial observability~\cite{zhang2021multiagent}.
Besides, we consider homogeneous agents, enabling shared actor parameters and more efficient learning through experience sharing.

\begin{figure*}
    \centering
    \includegraphics[scale=0.9]{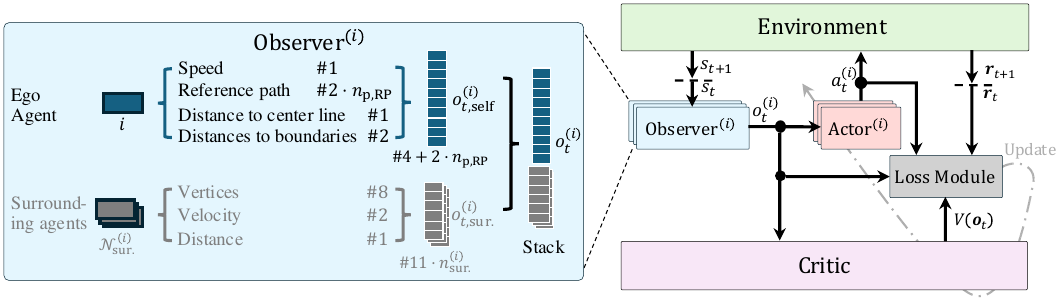}  
    \caption{
        Overview of the proposed decentralized \ac{marl} framework \texttt{SigmaRL}.
        $\timeStep$: time step;
        $\anAgent \in \{1,\dots,N\}$: agent index.
    }\label{fig:frameworkOverview}
\end{figure*}

\Cref{fig:frameworkOverview} overviews our decentralized framework aiming to solve \cref{prob:generalizableAgents}.
At time step $\timeStep$, each agent $\anAgent$ receives its partial observation $\observation\ofAnAgent\ofATimeStep$ from its designated observer, $\text{Observer}\ofAnAgent$, and executes its policy via its actor, $\text{Actor}\ofAnAgent$, to generate action $\action\ofAnAgent\ofATimeStep$. 
Subsequently, the environment updates its state based on the joint actions $\jointActions\ofATimeStep$ of all agents.
The critic, serving as a state-value function, estimates the potential future reward $\valueFunction(\jointObservations\ofATimeStep)$ based on the joint observations of all agents $\jointObservations\ofATimeStep$.
The loss module updates the critic and actor, details of which are referred to~\cite{schulman2017proximal,lowe2017multiagent}.

\subsection{RL Environment}\label{sec:rlEnv}
We use VMAS~\cite{bettini2024vmas}, a vectorized multi-agent simulator for collective robot learning, as our RL environment framework.
We customize the environment to align with our \ac{cpmlab}~\cite{kloock2021cyberphysicalc}, an open-source testbed for \acp{cav}.

\begin{figure}[t!]
    \centering
    \includegraphics[scale=0.7]{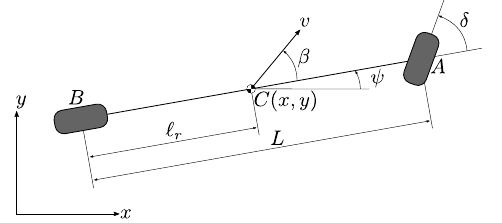}
    \caption{
        Kinematic single-track model. 
        $C$: \ac{cg};
        $x, y$: $x$- and $y$-coordinates;
        $v$: velocity;
        $\beta$: slide slip angle;
        $\psi$: yaw angle;
        $\delta$: steering angle;
        $L$: wheelbase.
    }\label{fig:bicycleModel}
\end{figure}

We use the nonlinear kinematic single-track model~\cite[Sec.~2.2]{rajamani2006vehicle}, visualized in \cref{fig:bicycleModel}, to model the agent dynamics.
It uses speed $v$ and steering angle $\delta$ as control actions.
We use continuous action spaces with $v \in [-0.8, 0.8]$ \si{\meter\per\second} and $\delta \in {[-35,35]}^{\circ}$.
We consider each agent a rectangle with a width of \SI{0.08}{\meter} and a length of \SI{0.16}{\meter}.

\subsection{Observation Design}\label{sec:observationDesign}
In this section, we propose five observation-design strategies to enhance the sample efficiency and the generalization of RL agents: the first one concerns the coordinate system used to represent surroundings, the second and third pertain to surrounding agents, and the last two are related to lanes. We verify the effectiveness of these strategies in the ablation studies in \cref{sec:ablationStudies}.

Most state-of-the-art RL agents for motion planning use image-like observations, which hold the information in an unstructured manner, usually requiring deep neural network architectures such as deep \acp{cnn} to extract relevant features.
However, using image-like observations hardens the learning process and often leads to large samples and time to converge~\cite{li2019reinforcement}, which contradicts the objective of this work.
Therefore, we use structured data with dense information to represent relevant information for motion planning, allowing the use of shallower neural networks for learning.
\begin{figure}[t!]
    \centering
    \includegraphics[scale=1.0]{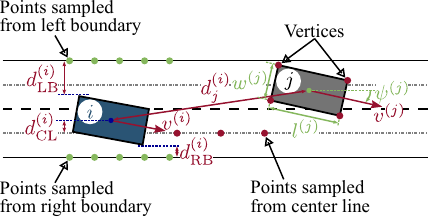}
    \caption{
        Observations of agent $\anAgent$. Red: efficient observation (ours). Green: inefficient observation (not ours). 
    }\label{fig:observations}
\end{figure}

\subsubsection{Use Ego View}
Mainly two types of presentations of surroundings are used in the motion planning of \acp{cav}: Ego view and bird-eye view (also known as top-down view).
While the bird-view presentation of surroundings is popular in traditional optimization-based methods, machine learning-based methods favor both.
Bird-view presentation is particularly advantageous in imitation learning for generating training data and for representing spatial relationships in a human-readable format~\cite{bansal2018chauffeurnet}.
However, we suggest using ego-view representation in RL, as it is naturally used in biological systems and thus adheres to the fact that most core algorithms of RL were inspired by biological learning systems~\cite[p. 4]{sutton2018reinforcement}.
The upcoming ablation studies in \cref{sec:ablationStudies} confirm its significant ability in enhancing sample efficiency and generalization.

\subsubsection{Vertices of Surrounding Agents}
Agents need to observe their surrounding agents to make risk-aware motion planning.
One traditional way is to observe their poses\footnote{
    We refer the pose of an agent to the combined position and orientation of this agent, where the position refers to the coordinates of its \acs{cg}.
} along with their geometric dimensions like lengths and widths, with the hope that agents implicitly learn the occupancy\footnote{
    We refer the occupancy of an agent to the spatial volume it occupies in the environment at any given time.
} of their surrounding agents.
However, we suggest explicitly observing that occupancy.
We propose to represent an agent's occupancy with the coordinates of its vertices.
See \cref{fig:observations} as an example, where the ego agent $\anAgent$ observes the vertices shown in red points of its surrounding agent $\anotherAgent$.
Our ablation studies in \cref{sec:ablationStudies} proves that this would notably lower the collision rate between agents, which we term agent-agent collision rate.

\subsubsection{Distances to Surrounding Agents}
Beyond observing the occupancy of surrounding agents, we suggest additionally observing the distances to them.
\Cref{fig:observations} depicts the ego agent $\anAgent$'s distance to its surrounding agent $\anotherAgent$, denoted as $\distance\ofAnAgent_{\anotherAgent}$.
Our ablation studies in \cref{sec:ablationStudies} show that this strategy would further reduce the agent-agent collision rate.

\subsubsection{Distances to Lane Boundaries}
Essentially, agents need to observe lane boundaries to prevent collisions with them (also called off-road events).
One traditional approach to represent a lane boundary is to discretize it with a polyline, such as a Lanelet~\cite{bender2014laneletsa,poggenhans2018lanelet2a}, and observe the coordinates of the points sampled from the polyline.
See \cref{fig:observations} as an example, where the sampled points of the lane boundaries of the ego agent $\anAgent$ are depicted by green points.
However, we remark that this approach, despite being widely used, is inefficient, since the number of sampled points may be large to preserve fidelity.
To counteract this shortcoming, we propose a more compact observation approach---observing the distances to lane boundaries.
As an example, see agent $\anAgent$'s distances to its left and right boundaries, denoted as $\distance\ofAnAgent_{\text{LB}}$ and $\distance\ofAnAgent_{\text{RB}}$, respectively.
Our ablation studies in \cref{sec:ablationStudies} reveal that this approach would remarkably reduce the collision rate between agents and lane boundaries, which we term agent-lane collision rate.

\subsubsection{Distances to Lane Center Lines}
The observation of lane center lines contributes to lane-following performance.
In this work, for each agent at each time step, we dynamically sample a fixed number of the most front points on its lane center line, serving as its short-term reference path.
See the three red points in front of agent $\anAgent$ in \cref{fig:observations} as an example.
We propose to let each agent observe its deviation from its lane center line, depicted exemplarily by $\distance\ofAnAgent_{\text{CL}}$ in \cref{fig:observations}, which denotes agent $\anAgent$'s distance to its lane center line.
Our ablation studies in \cref{sec:ablationStudies} demonstrates that this would greatly increase agents' lane-following performance.

Except for the above five observation-design strategies, we let each agent observe its own speed and the relative velocities of its surrounding agents. 
Adhering to partial observability, we allow each agent to observe only a limited number of surrounding agents.

On the left side, \cref{fig:frameworkOverview} overviews each agent $\anAgent$'s observation.
At each time step $\timeStep$, agent $\anAgent$'s observation consists of two parts: self-observation $\observation\ofAnAgent_{\timeStep,\text{self}}$ and the observation of surrounding agents $\observation\ofAnAgent_{\timeStep,\text{sur.}}$.
Specifically, the self-observation $\observation\ofAnAgent_{\timeStep,\text{self}}$ consists of its own speed, a short-term reference path sampled from its lane center line, distance to this center line, and distances to its left and right lane boundaries.
This results in a total of $4 + 2 \cdot \numOfPointsRef$ data points, where $\numOfPointsRef$ denotes the number of points building the short-term reference path.
Let $\setOfAgents\ofAnAgent_{\timeStep,\text{sur.}}$ denote the set of surrounding agents observable by agent $\anAgent$.
The observation of each agent $\anotherAgent \in \setOfAgents\ofAnAgent_{\timeStep,\text{sur.}}$, $\observation\ofAnAgent_{\timeStep\ofAnotherAgent}$, consists of agent $\anotherAgent$'s vertices, velocity, and the distance to agent $\anAgent$, totaling to eleven data points.
Consequently, this yields the observation of surrounding agents at time step $\timeStep$:    $\observation\ofAnAgent_{\timeStep,\text{sur.}} \coloneq \bigcup_{\anotherAgent \in \setOfAgents\ofAnAgent_{\timeStep,\text{sur.}}} \observation\ofAnAgent_{\timeStep\ofAnotherAgent}$.
Agent $\anAgent$ stacks the above two parts to form its final observation at time step $\timeStep$:
$\observation\ofAnAgent\ofATimeStep \coloneq \observation\ofAnAgent_{\timeStep,\text{self}} \cup \observation\ofAnAgent_{\timeStep,\text{sur.}}$.
Denoting further by $\numOfObservedSurroundingAgents\ofAnAgent \coloneq \left\lvert \setOfAgents\ofAnAgent_{\timeStep} \right\rvert$ the number of observed surrounding agents yields the total observation size of each agent $\anAgent$:
$
    \left\lvert \observation\ofAnAgent\ofATimeStep \right\rvert =
    4 + 2 \cdot \numOfPointsRef + 
    11\cdot\numOfObservedSurroundingAgents\ofAnAgent.
$

\begin{remark}[Generalization and Sample Efficiency]
    The proposed dense representation of observations crafts general yet crucial features for motion planning, such as vertices of surrounding agents and distances to them, as well as distances to lane boundaries and center lines.
    These features are broadly applicable across nearly all traffic scenarios, which grants RL agents a strong generalization potential to handle unseen scenarios.
    This way, we are allowed to train them on a challenging scenario to effectively learn a generalizable policy, without overfitting to specific scenes.
    Moreover, unlike end-to-end motion planning methods that demand sophisticated feature extractors, this approach densifies the information in observations to facilitate the learning process, thereby enhancing sample efficiency.
\end{remark}

\begin{remark}[Practicability]
    The proposed observation-design strategies necessitate observing the distances to lane boundaries, lane center lines, and distances to lane center lines.
    This poses no issue if a high-definition map representing lane boundaries and center lines exists and if agents can localize themselves within it, making it possible to calculate this information.
    Besides, the proposed strategies also necessitate observing the vertices of surrounding agents and the distances to them, which can be realized through communication. 
    Given that agents can localize themselves within the map, they know their poses and can calculate the coordinates of their vertices based on their geometric dimensions. They can then communicate these coordinates to other agents that need this information.
\end{remark}

\section{Experiments}\label{sec:ablationStudies}
In this section, we conduct five ablation studies to validate the efficacy of the observation-design strategies proposed in \cref{sec:observationDesign} and use four unseen scenarios to test our RL agents' generalization.
We present the experiment setups in \cref{sec:ExpSetups} and discuss the experiment results in \cref{sec:ablationStudies}.
We open-source our repository\footnote{
    \href{https://github.com/bassamlab/SigmaRL}{\small github.com/bassamlab/SigmaRL}
}
for reproducibility and provide a video therein to demonstrate our experiments.

\subsection{Training and Testing Setups}\label{sec:ExpSetups}
In our multi-agent \ac{ppo}, the centralized critic has four layers, one input layer, two hidden layers, and one output layer, with each hidden layer having 256 nodes.
We use \texttt{Tanh} as the activation function.
The actor has the same neural network architecture as the critic.
We use a sample time of \SI{50}{\milli\second} for both training and testing.
We set the number of observed surrounding agents $\numOfObservedSurroundingAgents$ to two and the number of points on a short-term reference path $\numOfPointsRef$ sampled from a lane center line to three.
We use a discount factor $\gamma$ of 0.99. 

We train \numberOfModels{} RL models, i.e., $\setOfModels = \{\model_0,\dots,\model_5\}$.
$\model_0$, called our model thereafter, incorporates all five observation-design strategies proposed in \cref{sec:observationDesign}.
From models $\model_1$ to $\model_5$, each model omits one of these strategies in order.
For instance, instead of using an ego view, $\model_1$ uses a bird-eye view.
Since our proposed observation-design strategies focus on general features that are applicable to most traffic scenarios, we are allowed to train the models only on the intersection of the CPM Scenario to learn generalizable policies.
Besides, we train each model with only four agents but test them with more agents.
We set the number of training episodes to 250, where 4096 samples are collected per episode, leading to approximately only one million samples.
We predefine several long-term reference paths and randomly select one for each agent.
Besides, we initialize the agents with random initial states.
Once a collision occurs, we reset all agents with random states and also with randomly selected reference paths from the predefined reference paths.
To ensure the feasibility of the initial states, we ensure the initial distances between agents are large enough (larger than 1.2 times the diagonal lengths of the agents).

\begin{figure}[t!]
    \centering
    \begin{subfigure}[b]{0.25\textwidth}
        \centering
        \includegraphics[scale=0.4]{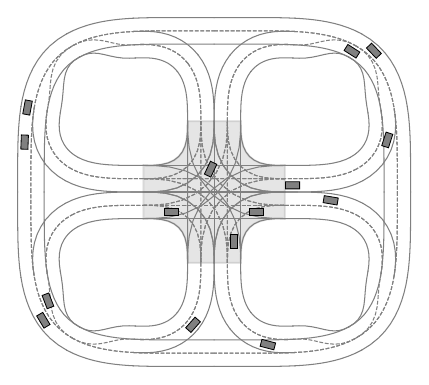}
        \caption{CPM Scenario.}
        \label{fig:sub-cpm}
    \end{subfigure}%
    \begin{subfigure}[b]{0.25\textwidth}
        \centering
        \includegraphics[scale=0.5]{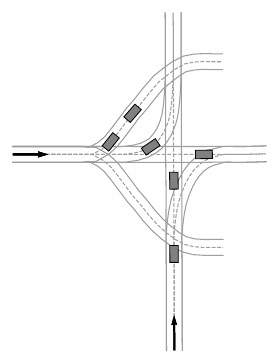}
        \caption{Intersection Scenario.}
        \label{fig:sub-intersection}
    \end{subfigure}

    \begin{subfigure}[b]{0.25\textwidth}
        \centering
        \includegraphics[scale=0.5]{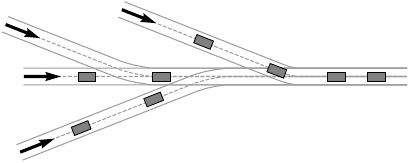}
        \caption{On-Ramp Scenario.}
        \label{fig:sub-onramp}
    \end{subfigure}%
    \begin{subfigure}[b]{0.25\textwidth}
        \centering
        \includegraphics[scale=0.5]{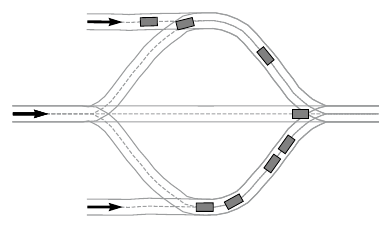}
        \caption{Roundabout Scenario.}
        \label{fig:sub-roundabout}
    \end{subfigure}

    \caption{Training and testing scenarios. Train only on the intersection of the CPM Scenario (see gray area). Test in all four scenarios, with the depicted numbers of agents.}
    \label{fig:allScenarios}
\end{figure}

We test the models in four completely unseen scenarios to evaluate the generalization of the learned policies: the entire CPM Scenario (\cref{fig:sub-cpm}) and three other scenarios handcrafted in Open Street Map~\cite{haklay2008openstreetmap}: an intersection (\cref{fig:sub-intersection}), an on-ramp (\cref{fig:sub-onramp}), and a roundabout (\cref{fig:sub-roundabout}), with 15, 6, 8, and 8 agents, respectively.
Since we train agents only on the intersection of the CPM Scenario, all four scenarios are unseen\footnote{
    Strictly speaking, the CPM Scenario is partially unseen, since its intersection is used for training.
} for them, which significantly challenges their generalization ability.
To obtain convincing results, we conduct 32 simulations in each scenario for each model, with each simulation having 1200 time steps.
Since the sample time is \SI{50}{\milli\second}, each simulation lasts for $\SI{50}{\milli\second} \times 1200 = \SI{1}{\minute}$.
In each simulation, we initialize the agents with random initial states.
Once a collision occurs, we reset only the colliding agents so that other agents will not be unnecessarily reset to ``safe'' states.

For each simulation and for each model, we compute three performance metrics: collision rate \collisionRate{}, center line deviation \centerLineDeviation{}, and average speed \averageSpeed{}, which are then used to compute the composite score \compositeScore{} via \cref{eq:score} to assess the model's sample efficiency defined in \cref{def:sampleEfficiency}.
Recall that we distinguish between agent-agent collision rate and agent-lane collision rate, denoted as \collisionRateAA{} and \collisionRateAL{}, respectively.
Summing them up yields the total collision rate $\collisionRate_{\text{total}} \coloneq \collisionRateAA + \collisionRateAL$.
We average the performance metrics over all simulations and show them in \cref{tab:results}.
Computing the composite score \compositeScore{} defined by \cref{eq:score} necessitates determining the weighting factors $w_1$, $w_2$, and $w_3$.
The weighting factors are expected to balance the relative importance and scale of each evaluation metric.
We treat each metric equally important and determine the weighting factors by inverting the average of the three performance metrics over all models, i.e.,
$w_1 = {|\setOfModels|} / {{\sum_{\forall \model_j \in \setOfModels} \collisionRate_{\text{total}, \model_j}}}$,
$w_2 = {|\setOfModels|} / {{\sum_{\forall \model_j \in \setOfModels} \centerLineDeviation_{\model_j}}}$, and
$w_3 = {|\setOfModels|} / {{\sum_{\forall \model_j \in \setOfModels} \averageSpeed_{\model_j}}}$,
where $|\setOfModels|$ denotes the number of models, which is \numberOfModels{} in our case.
This way, we balance the scale of the three performance metrics.
In summary, the composite score\footnote{Note that a composite score has no unit, and a higher value indicates better performance.} $\compositeScore_{\model_i}$ of model $\model_{i \in \{0,\dots,\maxModelIndex\}}$, is calculated as 

{\scriptsize
\begin{equation}\label{eq:scoreNew}
    \compositeScore_{\model_i} = 
    -\frac{|\setOfModels| \cdot \collisionRate_{\model_i}}{{\displaystyle\sum_{\forall \model_j \in \setOfModels} \collisionRate_{\text{total}, \model_j}}} 
    - \frac{|\setOfModels| \cdot \centerLineDeviation_{\model_i}}{{\displaystyle\sum_{\forall \model_j \in \setOfModels} \centerLineDeviation_{\model_j}}}
    + \frac{|\setOfModels| \cdot \averageSpeed_{\model_i}}{{\displaystyle\sum_{\forall \model_j \in \setOfModels} \averageSpeed_{\model_j}}}.
\end{equation}
}

\subsection{Results and Discussions}\label{sec:results}
\begin{figure}[t!]
    \centering
    \includegraphics[scale=0.9]{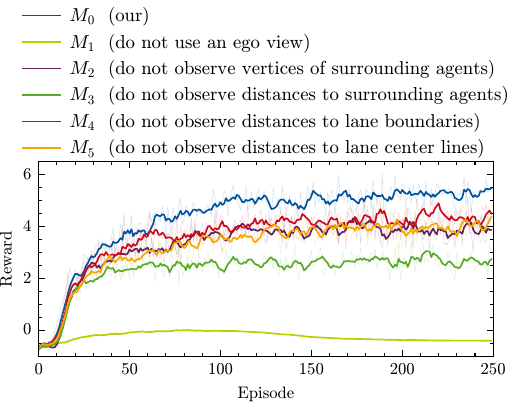}
    \caption{
        Mean reward per episode when training the \numberOfModels{} models $\model_{i \in \{0,\dots,\maxModelIndex\}}$.
        Model $\model_0$ incorporates all five observation-design strategies we proposed in \cref{sec:observationDesign}, whereas models $\model_1$ to $\model_5$ each omit one of these strategies.
    }\label{fig:reward}
\end{figure}

\Cref{fig:reward} depicts the mean reward per episode of the \numberOfModels{} models during training.
Due to the dense representation of observations, training each model takes less than one hour on a single CPU (Apple M2 pro with 16 GB of RAM, less than \SI{15}{\percent} CPU utilization).
Notably, our model $\model_0$, which incorporates all five observation-design strategies, exhibits the fastest learning speed and the highest episode reward.
In contrast, model $\model_1$, utilizing a bird-eye view rather than an ego view, demonstrates the lowest learning efficiency.
Moreover, model $\model_3$, which omits the third observation-design strategy---observing distances to surrounding agents, also shows lower learning efficiency. 
Models $\model_2$, $\model_3$, and $\model_5$ have similar learning curves, suggesting that the second, third, and fifth observation-design strategies contribute similarly to the learning process.

\Cref{tab:results} lists the testing results for all \numberOfModels{} models in four scenarios depicted in \cref{fig:allScenarios}.
We highlight the best performance metric in bold and the best composite score with an ellipse for each row.
Model $\model_1$, employing a bird-eye view, acts overly conservatively in the Intersection, On-Ramp, and Roundabout Scenarios, as evidenced by its nearly zero average speed \averageSpeed{}.
This conservativeness understandably yields a collision rate close to zero.
Therefore, we exclude their performance metrics in these three scenarios from the candidates of the best values.

\begin{table}[t!]
    \centering
    \caption{
    Testing results of the \numberOfModels{} models in four scenarios.
    \collisionRateAA: agent-agent collision rate;
    \collisionRateAL: agent-lane collision rate;
    \collisionRateTotal: total collision rate;
    \centerLineDeviation: center line deviation;
    \averageSpeed: average speed;
    \compositeScore: composite score calculated by \cref{eq:scoreNew}.
    }
    \label{tab:results}
    \resizebox{0.46\textwidth}{!}{
    \begin{tabular}
    {
        @{}
        p{0cm}
        l
        S[table-format=1.1]
        S[table-format=1.2]
        S[table-format=1.2]
        S[table-format=1.2]
        S[table-format=1.2]
        S[table-format=1.2]
        S[table-format=1.2]
        @{}
    }
    \toprule
     & & & {$\model_0$} & {$\model_1$} & {$\model_2$} & {$\model_3$} & {$\model_4$} & {$\model_5$} \\
    \midrule
    \multirow{6}{*}{\rotatebox[origin=c]{90}{CPM Scenario}}
    & & {$\collisionRate_{\text{A-A}} \ [\si{\percent}]$} & \textbf{0.04} & 4.14 & 0.56 & 0.92 & 0.05 & 0.62 \\
    & & {$\collisionRate_{\text{A-L}} \ [\si{\percent}]$} & 0.35 & 21.83 & \textbf{0.02} & \textbf{0.01} & 0.52 & \textbf{0.01} \\
    & & {$\collisionRate_{\text{total}} \ [\si{\percent}]$} & \textbf{0.38} & 25.97 & 0.58 & 0.93 & 0.57 & 0.63 \\
    & & {$\centerLineDeviation{} \ [\si{\centi\meter}]$} & 5.18 & 16.03 & 4.50 & \textbf{4.28} & 4.60 & 5.06 \\
    & & {$\averageSpeed{} \ [\si{\meter\per\second}]$} & \textbf{0.74} & 0.43 & 0.69 & 0.72 & 0.73 & 0.72 \\
    \cmidrule(l){3-9}
    & & {\compositeScore} & 0.24 & -7.15 & 0.23 & 0.23 & {\ellipsed{\textbf{0.28}}} & 0.17 \\
    \midrule
    \multirow{6}{*}{\rotatebox[origin=c]{90}{Intersection}}
    & & {$\collisionRate_{\text{A-A}} \ [\si{\percent}]$} & \textbf{0.10} & \sout{0.02} & 1.33 & 2.42 & 0.88 & 1.76 \\
    & & {$\collisionRate_{\text{A-L}} \ [\si{\percent}]$} & 0.86 & \sout{0.20} & \textbf{0.03} & 1.73 & 0.47 & 1.25 \\
    & & {$\collisionRate_{\text{total}} \ [\si{\percent}]$} & \textbf{0.96} & \sout{0.22} & 1.35 & 4.16 & 1.35 & 3.01 \\
    & & {$\centerLineDeviation{} \ [\si{\centi\meter}]$} & 2.76 & \sout{2.44} & 2.60 & 3.64 & \textbf{2.47} & 3.59 \\
    & & {$\averageSpeed{} \ [\si{\meter\per\second}]$} & 0.71 & \sout{0.07} & 0.70 & 0.72 & 0.70 & \textbf{0.74} \\
    \cmidrule(l){3-9}
    & & {\compositeScore} & {\ellipsed{\textbf{-0.30}}} & -0.85 & -0.47 & -2.32 & -0.42 & -1.64 \\
    \midrule
    \multirow{6}{*}{\rotatebox[origin=c]{90}{On-Ramp}}
    & & {$\collisionRate_{\text{A-A}} \ [\si{\percent}]$} & \textbf{0.09} & \sout{0.01} & 0.55 & 3.56 & 0.49 & 2.56 \\
    & & {$\collisionRate_{\text{A-L}} \ [\si{\percent}]$} & \textbf{0.00} & \sout{0.01} & \textbf{0.00} & \textbf{0.00} & \textbf{0.00} & \textbf{0.00} \\
    & & {$\collisionRate_{\text{total}} \ [\si{\percent}]$} & \textbf{0.09} & \sout{0.03} & 0.55 & 3.56 & 0.49 & 2.56 \\
    & & {$\centerLineDeviation{} \ [\si{\centi\meter}]$} & \textbf{2.00} & \sout{2.38} & 2.16 & 4.15 & 2.44 & 3.74 \\
    & & {$\averageSpeed{} \ [\si{\meter\per\second}]$} & 0.69 & \sout{0.06} & 0.68 & 0.71 & 0.68 & \textbf{0.74} \\
    \cmidrule(l){3-9}
    & & {\compositeScore} & {\ellipsed{\textbf{0.38}}} & -0.77 & -0.08 & -3.21 & -0.13 & -2.19 \\
    \midrule
    \multirow{6}{*}{\rotatebox[origin=c]{90}{Roundabout}}
    & & {$\collisionRate_{\text{A-A}} \ [\si{\percent}]$} & \textbf{0.26} & \sout{0.09} & 2.10 & 4.78 & 1.20 & 3.59 \\
    & & {$\collisionRate_{\text{A-L}} \ [\si{\percent}]$} & 0.07 & \sout{1.21} & \textbf{0.00} & 0.37 & 0.06 & 0.28 \\
    & & {$\collisionRate_{\text{total}} \ [\si{\percent}]$} & \textbf{0.33} & \sout{1.30} & 2.10 & 5.15 & 1.26 & 3.87 \\
    & & {$\centerLineDeviation{} \ [\si{\centi\meter}]$} & \textbf{2.51} & \sout{2.24} & 2.44 & 4.05 & 2.47 & 3.75 \\
    & & {$\averageSpeed{} \ [\si{\meter\per\second}]$} & 0.67 & \sout{0.07} & 0.65 & 0.70 & 0.66 & \textbf{0.72} \\
    \cmidrule(l){3-9}
    & & {\compositeScore} & {\ellipsed{\textbf{0.15}}} & -1.21 & -0.61 & -2.38 & -0.25 & -1.69 \\
    \bottomrule
    \end{tabular}
    }
\end{table}

Remarkably, despite being trained exclusively on the intersection of the CPM Scenario, our model $\model_0$ demonstrates robust performance in the unseen scenarios.
This achievement confirms that our proposed observation-design strategies successfully grant the RL agents the capability to zero-shot generalize to unseen scenarios.
The agents achieve a collision rate of less than \SI{1.0}{\percent} while maintaining high traffic efficiency in the testing scenarios.
We gauge traffic efficiency using the performance metric \texttt{average speed} \averageSpeed{}. 
Given the maximum speed being set to \SI{0.8}{\meter\per\second}, our agents achieve average speeds of more than \SI{80}{\percent} of this maximum in the testing scenarios.
Note that traffic density---specifically, the number of agents---significantly influences traffic efficiency.
For intuition, we refer readers to the video within our open-source repository.

Overall, our model $\model_0$ outperforms the other five models.
It achieves the majority of the best values across the three performance metrics and the composite score.
Moreover, it secures the highest composite scores in the three handcrafted scenarios, i.e., the Intersection, On-Ramp, and Roundabout Scenarios.
Although it does not achieve the highest composite score in the CPM Scenario, its performance remains close to the model with the best composite score, scoring 0.24 versus 0.28 by model $\model_4$).
Owing to the omission of one of the proposed observation-design strategies, other five models underperform in some performance metrics:
\begin{itemize}
    \item Model $\model_1$, which uses a bird-eye view instead of an ego view, underperforms in almost all performance metrics, likely owing to low learning efficiency during training.
    \item Model $\model_2$, which observes surrounding agents' poses and geometric dimensions instead of their vertices, notably increases the agent-agent collision rate. Interestingly, it lowers the agent-lane collision rate, presumably because the ineffective observation of surrounding agents leads to an attention shift from surrounding agents to lanes.
    \item Model $\model_3$, which does not observe the distances to surrounding agents, suffers from the highest agent-agent collision rate in most scenarios. This highlights the critical role of distance observation in learning risk awareness to prevent collisions with other agents.
    \item Model $\model_4$, which observes the sampled points from the boundaries rather than the distances to them, excels in avoiding agent-lane collisions. However, it leads to a high agent-agent collision rate, possibly because it causes agents to overly focus on lane boundaries at the expense of neglecting their surrounding agents.
    \item Model $\model_5$'s omission of observing distances to center lines results in poor lane-following performance compared to model $\model_0$. It increases the average speed in the second, third, and fourth scenarios, primarily by compromising safety.
\end{itemize}

In summary, our experiments underline the effectiveness of the proposed observation-design strategy in enhancing sample efficiency and generalization.

\subsection{Limitations of Our Study}
The composite score calculated through \cref{eq:scoreNew} favors conservative models.
The most conservative model, which lets all agents stay stationary in a scenario, would get a composite score of zero, since all three performance metrics would be zero. Consequently, this score might misleadingly indicate better performance than other models, who may get negative scores if the scenario is challenging enough.

\section{Conclusions}\label{sec:conclusions}
In this paper, we presented our open-source \texttt{SigmaRL}, a \underline{s}ample-eff\underline{i}cient and \underline{g}eneralizable \underline{m}ulti-\underline{a}gent \underline{RL} framework for motion planning of \acp{cav}.
We formulated the motion planning problem as a partially observable Markov game and explored the under-explored area of how observation design affects sample efficiency and generalization. As outcomes, we proposed five strategies for designing information-dense, structured observations that enhanced both the sample efficiency and the generalization of RL agents.
These strategies focused on extracting general features applicable across various traffic scenarios.
In our numerical experiments, we required only one million samples and less than one hour on a single CPU to train our agents, suggesting outstanding sample efficiency.
Despite being only trained on an intersection, they demonstrated outstanding zero-shot generalization to completely unseen scenarios, including a new intersection, an on-ramp, and a roundabout.
Our results suggested that our observation-design strategies may be a viable approach toward achieving sample efficient and generalizable \ac{marl} for motion planning of \acp{cav}.

Future work will include comparing our proposed observation design with image-based observations regarding sample efficiency and generalization.

\bibliographystyle{IEEEtran}
\bibliography{main.bib}

\end{document}